\documentclass[10pt,twocolumn,letterpaper]{article}

\usepackage{cvpr}
\usepackage{times}
\usepackage{epsfig}
\usepackage{graphicx}
\usepackage{amsmath}
\usepackage{amssymb}

\usepackage[pagebackref=true,breaklinks=true,letterpaper=true,colorlinks,bookmarks=false]{hyperref}

\cvprfinalcopy 

\def\cvprPaperID{****} 

\ifcvprfinal\fi
\begin{document}

\title{Large-scale Video Classification guided by Batch Normalized LSTM Translator}

\author{Jae Hyeon Yoo\\
Mobile Communications Business\\
Samsung Electronics\\
{\tt\small {jaehyeon.yoo}@samsung.com}
}

\maketitle

\begin{abstract}
   Youtube-8M dataset enhances the development of large-scale video recognition technology as ImageNet dataset has encouraged
   image classification, recognition and detection of artificial intelligence fields.
   For this large video dataset, it is a challenging task to classify a huge amount of multi-labels. By change of perspective,
   we propose a novel method by regarding labels as words. In details, we describe online learning approaches to multi-label video
   classification that are guided by deep recurrent neural networks for video to sentence translator. We designed the translator
   based on LSTMs and found out that a stochastic gating before the input of each LSTM cell can help us to design the structural
   details. In addition, we adopted batch normalizations into our models to improve our LSTM models. Since our models are feature extractors,
   they can be used with other classifiers. Finally we report improved validation results of our models on large-scale Youtube-8M datasets and
   discussions for the further improvement.
\end{abstract}

\section{Introduction}

Video recognition technology is very important in the field of artificial intelligence. It is a challenging task because understanding context of a given video is related to high-level temporal causal relationship among the scenes. In addition, this technology can be applied to a variety of fields such as learning activity recognition or scene understanding in videos\cite{DBLP:journals/corr/DonahueHGRVSD14,DBLP:journals/pami/KarpathyF17}, detecting future incidents or criminals by tracking real-time CCTV videos\cite{chen2011face,sankaranarayanan2008object} and decoding cognitive thinking process of subject by analyzing temporal patterns of brain activity in fMRI images.\cite{kamitani2005decoding,kay2008identifying,norman2006beyond}

Fundamentally, video recognition technology is required to understand the topic of a given video. A video consists of the sequence of images that are correlated with each other. In the problem of tagging what information the video contains, it is possible to tag one identical label for multiple frames or tag multiple labels in one frame and not to tag for the rest of the frames. That is, the frames that contain the topics of the video are determined with respect to the distribution and relation among entire images in the video. The distribution of labels tagged in these frames of one video is different from that of other video. Moreover, the number of frames in videos varies, and the distribution is variable and unknown. In addition, multi-label classification problem can be solved by logistic regression, mixture model, SVM, but if dataset to be analyzed is large-scale, batch learning could not be applied and online learning should be considered.\cite{DBLP:journals/corr/Abu-El-HaijaKLN16}
Despite these challenging factors, we try to approach this classification problem from different point of view. If we view the label as a word, classifying multiple labels from a video can eventually be turned into a video to sentence translation, or video description problem. Recent advances that generate a scene description from a video can be applied to this problem as it is; recent papers have improved the quality of video description technology with the development of neural networks and the powerful combination of CNN and LSTM. We also use the LSTM decoder and transfer learning based on the mean pooling of CNN features. Here, for easy transfer learning, we use Youtube-8M dataset\cite{Google2017} because it already stores and provides Inception CNN visual features for each frame. Therefore, we focus on what better LSTM structure is and how to improve its generalization performance by using recent optimization trend called batch normalization.

The contributions of this paper are the following:
\begin{quote}
- We suggest an insight that multi-label classification can be transformed to the problem in video description framework and establish base LSTM model. And we explore different structures of LSTM-based feature extractor. 

- We investigate how to improve generalization of LSTMs by using batch normalization. We deal with issues that occur when we use BNLSTM as video description translator, such as feedback selection issue. We introduce stochastic gating mechanism to alleviate this issue and determine which structure is better for feedback loop in the feature extractor.

- Finally, we report validation results of our models on large-scale Youtube-8M datasets

\end{quote}

\begin{figure*}
\begin{center}
\includegraphics[width=0.95\linewidth]{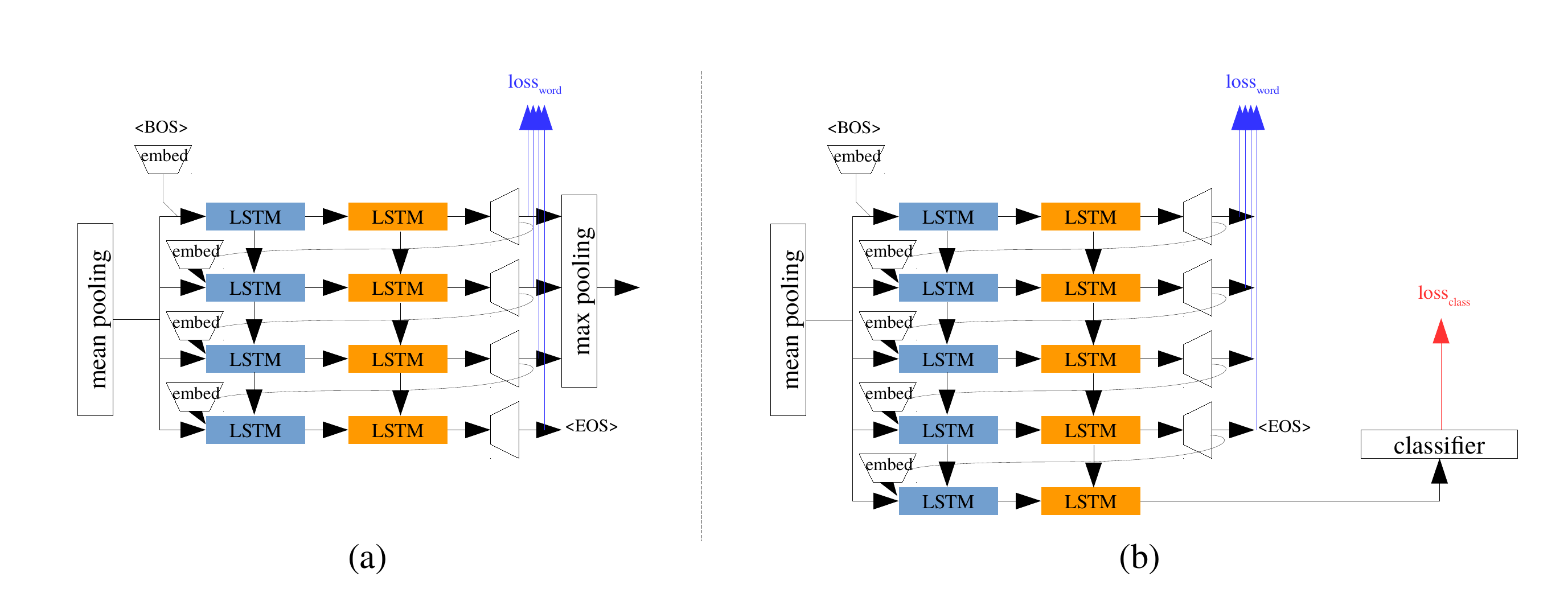}
\end{center}
   \caption{Detailed illustration of our LSTM models for video classification. (a) our base LSTM model, (b) as a variant of (a), the guided LSTM is designed for feature extractor to be used with the following classifier.}
\label{fig:short}
\end{figure*}

\section{Related Work}

  Donahue et al.\cite{DBLP:journals/corr/DonahueHGRVSD14} showed that a combination of CNN and LSTM can efficiently perform image caption, video description and activity recognition and the model can learn spatial and temporal compositional representations. Venugopalan et al.\cite{DBLP:journals/corr/VenugopalanXDRMS14} solved translating video to natural language problem by transfer learing from CNN structure, and performed LSTM decoding process after mean pooling those CNN features. These video frames also can be compressed into one visual feature vector by LSTM-based encoding process(Venugoplan et al.\cite{DBLP:journals/corr/VenugopalanRDMD15}) or 3D-CNN based representation(Yao et al. \cite{Yao15}). On top of LSTM Encoder-Decoder models, Cho et al.\cite{DBLP:journals/corr/ChoCB15} and Xu et al.\cite{xu2015show} extended the model for visual attentional framework and showed improved performance. It turns out to be useful using not only attentional mechanism but also features from additional information such as scoring some object classes or optical flow(Rohrbach et al.\cite{DBLP:journals/corr/RohrbachRS15}).
  To improve the validation performance of LSTM model, batch normalization method has been applied to LSTM models. Batch normalization uses batch mean and variance of input features for standardization to reduce internal covariate shift issue(Ioffe et al.\cite{DBLP:journals/corr/IoffeS15}). This batch normalization method is powerful and has recently become a trend, because this enables faster learning than dropout, preserving good generalization performance. Laurent et al.\cite{Laurent15} showed that the batch-normalized input-to-hidden transitions can lead to a faster convergence, and Cooijmans et al.\cite{DBLP:journals/corr/CooijmansBLC16} proposed a total reparameterization of LSTM by adding the hidden-to-hidden transitions, which improved generalization.
  
\section{Approach}

We propose a feature extractor for video classification guided by video description structure.
In general, neural machine translation finds patterns mapping between the input sentence of one natural language to the output sentence of another language. This idea has been effectively applied to the field of video description, because the input can be generalized into the forms of sequence of any features including video frames. 
In our model, we extend the original classification problem into the concept of video description.
By this change of perspective, we view each target label vector as a set of meaningful words, ``a sentence''. This idea results into the perspective that we can perform a translation from a video to a sequence. During translation process, the feature extractor can obtain aggregated features which are distinct to other sentence labels. This feature extraction process by translation is called ``guidance''. We can expect the final output of guidance can be utilized for video classification.

\subsection{Common structure}

There is a mean pooling layer to aggregate frame-level visual features and the output video-level features are input into all LSTM cell inputs.
To calculate ${loss}_{word}$, we split the learning target label into a set of one-hot vectors, and make a semantic word vector with embedding layer.
This word vector is also concatenated with the visual feature for input of LSTM cells. For guidance process,  semantic vectors for the virtual $<$BOS$>$ and $<$EOS$>$ tokens are introduced together.

\subsection{Basic LSTM structure for guidance}
The Long Short Term Memory\cite{hochreiter1997long} is one of the state-of-the-art Recurrent Neural Network that has been applied in neural machine translation\cite{johnson2016google}, image captioning\cite{DBLP:journals/corr/VinyalsTBE14}, video description\cite{DBLP:journals/corr/RohrbachRS15}, etc. LSTM deals with memorizing not only patterns observed until current time $t$, but also patterns of how to recall and forget correlations throughout the patterns based on hidden states $h_t$, internal memory cell state $c_t$ and three gates $i_t$, $o_t$, $f_t$. $g_t$ is a candidate memory cell state from the current input and the previous hidden:
\begin{subequations} \label{eq:lstm_model}
\begin{align}
i_t &= \sigma (W^i x_t \oplus w_t + U^i h_{t-1} + b_i) \\
o_t &= \sigma (W^o x_t \oplus w_t + U^o h_{t-1} + b_o) \\
f_t &= \sigma (W^f x_t \oplus w_t + U^f h_{t-1} + b_f) \\
g_t &= \tanh (W^g x_t \oplus w_t + U^g h_{t-1} + b_g) \\
c_t &= f_t\odot  c_{t-1} + i_t \odot g_t \\
h_t &= o_t \odot \tanh c_t 
\end{align}
\end{subequations}
  where $\oplus$ is a vector concatenation operator, $\odot$ is the element-wise multiplication between two vectors, W's are weight matrices from input to hidden states, U's are weight matrices from hidden to hidden. All weight matrices and biases b's are model parameters to be trained.
The input is composed of two parts. The first part can be any form of comprehensive feature that represents the whole frames of a given video.
Here, we set the mean pooled frame featrue as input of our model, including video and audio components of each frame. YouTube-8M dataset provides it as ``video-level'' feature. The second part is word embedding vector for guidance process. For any given instance, we split one multi-label target vector into many one-hot word vectors  $(y_1, ... , y_T)$, where $T$ is the number of tags in the target label vector. Finally we add embedding layer to squeeze the high dimensional sparse vectors into the lower dimensional dense word vectors  $(w_1, ... , w_T)$. Then the averaged frame feature $x$
is duplicated and concatenated with word vectors $w_t$, finally input to LSTM model at each time step, as $(x_1 \oplus w_1, ... , x_T \oplus w_T)$.  The intermeidate hiddens $(h_1, ... , h_T)$ are outputs of LSTM cells in charge of guiding memory of LSTM converging to the final goal state. These outputs are projected back into high dimensional space to get a distribution over all of the words in the vocabulary. Then, for each step, our LSTM models estimate conditional probability :
\begin{equation}
P(y_t, ... , y_1 | x_t \oplus w_t, ... ,x_1 \oplus w_1) = \prod_{1 \le t \le T}{P(y_t | h_{t-1})}
\end{equation}
and maximizes cross entropy of each word. After all, The final hidden state $h_T$ is used for video classification.
We can have benefits from this change of viewpoint in terms of classification performance as well as learning time. Our target dataset, YouTube-8M, contains videos with at most 300 frames annotated by 3.4 labels in average, maximally around 30 labels. Therefore, searching for the features in guidance process takes only about $\frac{1}{10}$ to $\frac{1}{100}$ times LSTM steps than learning in time domain. 
As depicted in (a) of figure \ref{fig:long}, each LSTM cell output passes through a common word projection layer which maps input vector into the original target word vector space.
Since this output has a meaning of likelihood distribution of the word vocabulary, we use softmax function as activation to make it a probabilistic distribution.
For training, we calculate cross entropy for each word vector and aggregate them. It means this LSTM structure is guided by word losses.
Besides, to calculate the overall output vector, max pooling layer aggregates all of the distribution outputs.

\subsection{LSTM as feature extractor}

Since the internal dynamics of LSTM is guided by sentence learning structure, it makes us to hypothesize that the final hidden state of LSTM is viewed
as a condensed feature including sentence inference path from $w_1$ to $w_T$. This idea makes us to design a different LSTM structure as a feature extrator,
which can make synergetic effect in collaboration with other classifiers. This design is illustrated in (b) in figure \ref{fig:long}.

\subsection{Stochastic Gating Mechanism}

\begin{figure}[t]
\begin{center}
   \includegraphics[width=0.95\linewidth]{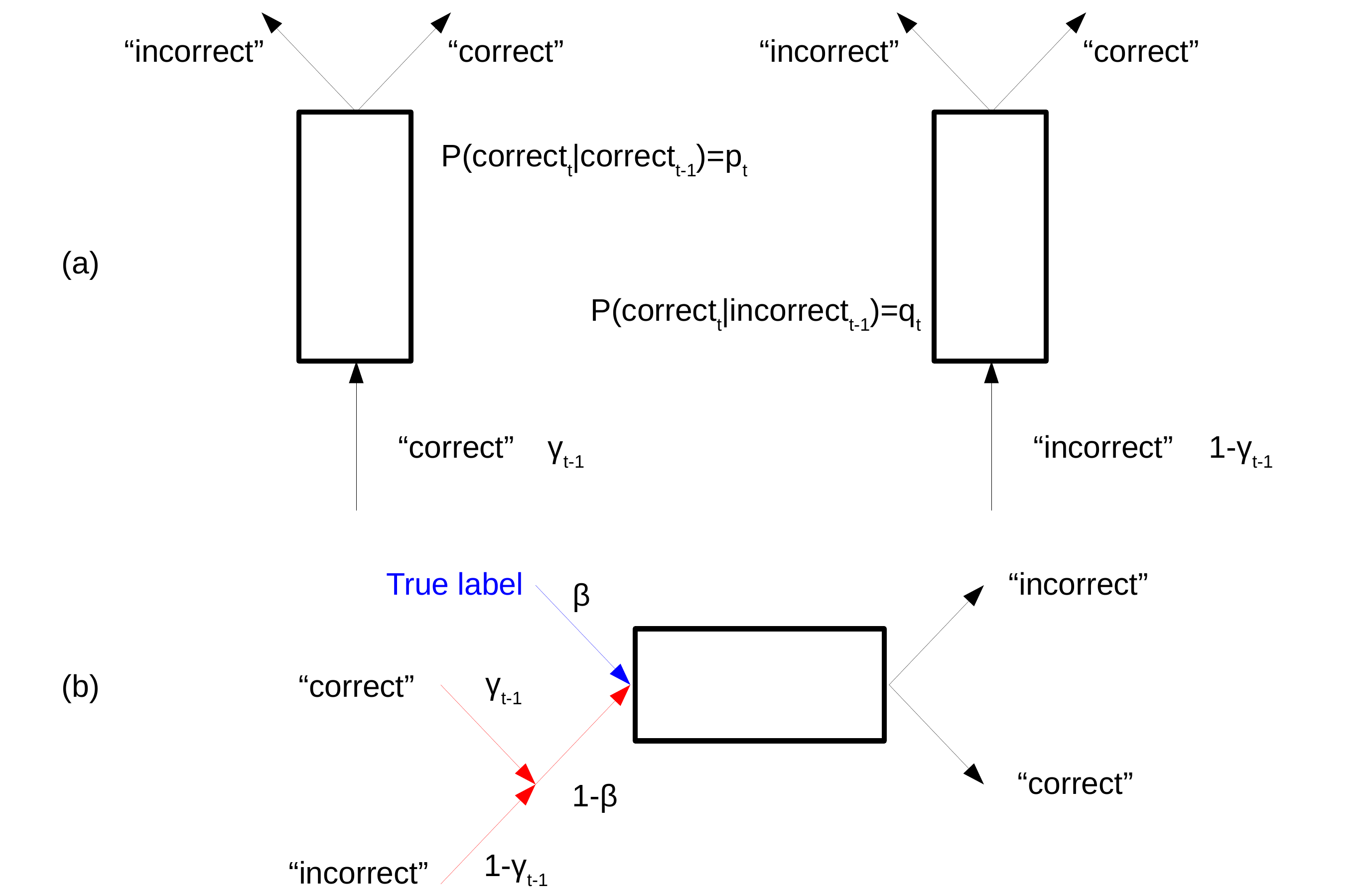}
\end{center}
   \caption{Asymptotic explanation about our stochastic gating mechanism which guides stacked LSTMs by teaching each LSTM cell 
   "label" during the intermeidate procedures. (a) This approach faces two possible cases: the input is correct or not. Therefore, our approach makes balance between these two cases by random process of gating. (b) The probability that gate will open to the ground truth label embedding vector at each time $t$ is $\beta$, which we call label injection probabililty. So, with probability $1-\beta$, Gates are open to 
   the output embedding vector of the previous cells. During inference process, the $\beta$ 
   is fixed to 0 and the model utilizes only the cell outputs of model itself. This method
   improves not only generalization performance but also the ascending speed of learning curve.
   }
\label{fig:long}
\label{fig:onecol}
\end{figure}

When it comes to input word vectors, many word generation structures utilized ground truth labels as input sequence for train phase.
It is switched to LSTM cell outputs when it performs inference. When we investigate this LSTM structure in detail, we face to a critical issue that
if we use ground truth label embedding vector as $w_t$, it actually leads our model to overfitting: For training phase, LSTM seems to learn
not the hidden patterns in $x_t$ but just $w_t$ itself and seems to bypass $w_1$ to $w_T$ to the final 
hidden state $h_T$. We questioned what the real effect of this switching is and how it is related to overfitting within our models. 
To figure out the cause, we added a stochastic gating before the input of each LSTM cell as one of structural variants. We can exploit both ground truth label and the embedding vector projected from the previous cell output as $w_t$ and this gate opens to ground truth labels with probability $\beta$ and to the previous cell output with $1-\beta$. By this structure we can avoid overfitting phenomena and consider which value of $\beta$ is helpful for learning. This stochastic gating mechansim is illustrated in Figure \ref{fig:short}. We can briefly touch the core concept of SG by using approximately simplified asymptotic model in Figure \ref{fig:long}. (a) if $\beta=1$, the model uses only ground truth labels as word vector $w_t$. This leads model to learn only $P({correct}_t | {correct}_{t-1})=p_t$ and there is no concern about $P({correct}_t | {incorrect}_{t-1})=q_t$, that affects lower generalization performance. (b) if $\beta=0$, the previous cell output is used for $w_t$. Let $\gamma_t$ be the probability that the tag at time $t$ is correct, 

\begin{subequations} 
\begin{align}
\gamma_t &= P({correct}_t | {correct}_{t-1}) P({correct}_{t-1}) \nonumber \\
        &+ P({correct}_t|{incorrect}_{t-1})P({incorrect}_{t-1}) \label{eq:1}\\
&=p_t \gamma_{t-1} + q_t (1-\gamma_{t-1})
\end{align}
\end{subequations}
If we assume that learning algorithm converges to an equilibirum state as time t goes to infinity
($\lim_{t\to \infty} \gamma_t = \gamma$, $\lim_{t\to \infty} p_t = p$, $\lim_{t\to \infty} q_t = q$),
\begin{subequations}
\begin{align}
\gamma &= p \gamma + q (1-\gamma) \nonumber \\
\gamma &= \gamma_0 = \frac{q}{1-p+q}
\end{align}
\end{subequations}

Now, (c) let us consider the case that $\beta \in (0,1)$. As the diagram is depicted, The probability that the input tag at time t is correct increases as ground truth label injection occurs with probability $\beta$.
\begin{subequations}
\begin{align}
P({correct}_{t-1}) &= \beta + (1-\beta) \gamma_{t-1} \nonumber \\
P({incorrect}_{t-1}) &= (1-\beta) (1-\gamma_{t-1}) \nonumber
\end{align}
\end{subequations}
 
 In this case, the probability of being correct at time t is computed as the equation \ref{eq:1}:
\begin{subequations}
\begin{align}
\gamma_t &= p_t (\beta + (1-\beta) \gamma_{t-1}) + q_t (1-\beta) (1-\gamma_{t-1}) \nonumber \\
\gamma &= \gamma (\beta) = \frac{p\beta + q(1-\beta)}{1 - (1-\beta)(p-q)}
\end{align}
\end{subequations}

Let us compare the $\gamma_0$ and $\gamma(\beta)$. If the learning algorithm trained the model to output the intermediate tags correctly with a high probability, it may be a good start to consider the case $p > q$ at first. Since the numerator is a weighted average between $p$ and $q$, it becomes larger than $q$. In addition, the negative term $-(p-q)$ of the denominator decreases by a factor of $1-\beta$, resulting into the increase of $\gamma(\beta)$. That is, if $p > q$, $\gamma_0 < \gamma(\beta)$ for $\beta > 0$, which means approximately it has a higher asymptotic limitation of learning curve than that of $\beta=0$ case. 

In other case, $p < q$ means the learning algorithm has trained the model to find patterns from the previous incorrect tag input to the correct tags and it is better than correct-to-correct tags. This case is possible if the amount of incorrect input tags are more trained than that of correct input tags. This results into the reversed relation: $\gamma_0 > \gamma(\beta)$. 

\subsection{Batch Normalized LSTM}

To improve our LSTM models, we adopted batch normalizations into our models. Firstly we just added two BN layers between classifier and LSTM final state and at the output word projection layer, respectively.
This structure doesn't modify LSTM itself. But the next model, Batch normalized LSTM (BNLSTM)\cite{DBLP:journals/corr/CooijmansBLC16,Laurent15} has interal batch normalization for reparameterization hiddens and cell memory:
\begin{subequations} \label{eq:bnlstm_model}
\begin{align}
\tilde{x}^j_t &= BatchNorm(W^j x_t \oplus w_t) \\
\tilde{h}^j_t &= BatchNorm(U^j h_{t-1}) \\
k_t &= \sigma ( \tilde{x}^j_t + \tilde{h}^j_t + b_j) \\
g_t &= \tanh ( \tilde{x}^g_t + \tilde{h}^g_t + b_g) \\
c_t &= f_t\odot  c_{t-1} + i_t \odot g_t \\
\tilde{c}_t &= BatchNorm(c_t) \\
h_t &= o_t \odot \tanh \tilde{c}_t
\end{align}
\end{subequations}
where $j=i,o,f,g$ and $k=i,o,f$

\section{Experimental Setup}

  This section illustrates the process of evalution for our approach. Firstly,
we explain about the Youtube-8M dataset that we worked on.
Secondly, we describe the evaluation metrics and lastly,
the implementation details of our models.

\subsection{YouTube-8M dataset}

YouTube-8M dataset\cite{Google2017} is a large-scale video benchmark dataset collected from Google YouTube.
It provides 8 Million video URLs with 4716 classes (video tags). Every video is 
tagged by 3.4 labels in average, and maximum number of labels in a video is around 30. 
For each video, there are two different levels; 
video-level and frame-level. It provides the videos 
as not pixel-level raw frames, but feature representation vectors extracted by 
Convolutional Neural Network(CNN) such as Inception network. That is, the dataset 
already has extracted significant feature vectors with 1024 dimension from videos 
by each frame per second. In addition, it also contains audio feature vectors 
with 128 dimension synchronized by video features. Each video has at most 300 frames, 
which consist of frame-level datasets, and one average pooled frame, which is video-level data.
They are all stored in froms of TensorFlow Record (tfrecord) binary files, to boost
up the loading and preprocessing speed. There are 4096 train tfrecords files, 
4096 validation files, and 4096 test files respectively. Frame-level dataset,
especially frame-level train dataset requires a huge amount of storage space, i.e. 1.2TB, 
Averaged pooled video-level (inception feature + audio feature) datasets are provided
due to the above reason. Video-level training dataset requires only less than 30 GB.
In this paper, we focus on video-level datasets to implement video classifier.
This is possible because recent studies \cite{DBLP:journals/corr/DonahueHGRVSD14,DBLP:journals/corr/VenugopalanXDRMS14} proved that mean pooling layer 
can be one of the efficient methods to aggregate frames in a video.

\subsection{Evaluation Metrics}

  For information retrieval, we can measure three different evaluation metrics
for the performance of topic classifiers such as Hit@k, PERR and GAP. \cite{DBLP:journals/corr/Abu-El-HaijaKLN16}
Hit@k is the fraction of retrieved samples that include one or more ground truth labels in top $k$ predictions
PERR means Precision at Equal Recall Rate, which measures the averaged fraction of how many predictions
are in the size of a set of ground truth labels, not just fixed value $k$. The calculation of both Hit@k and PERR
are based on ranking entity(label) scores from predictions. Finally, GAP is from the concept of averaged precision.
This GAP is a standard evaluation for YouTube-8M dataset\cite{Kaggle2017}. The detailed definitions of these
metrics can be found in \cite{DBLP:journals/corr/Abu-El-HaijaKLN16, Google2017}. Especially, \cite{Google2017} provides automatic evaluation tools for these metrics,
so we use them for this experiments.

\subsection{Experimental details of our models}

\textbf{Baseline Description} For multi-label video classification on Youtube-8M dataset,
logistic classifier and mixture of experts model are applied for video-level classification\cite{DBLP:journals/corr/Abu-El-HaijaKLN16}.
We also use them as our baseline models provided in starter code published by google\cite{Google2017}.
Since our model can be unified with classifiers including these baseline models, we can improve our models by boosting up 
the classifiers by adding dropout layers or extending dimension of layers, etc.
In this paper, we don't focus on these classifiers and leave them for the further work.

\textbf{Base LSTM model}
We implemented 2 layered standard LSTMs as explained in Section 3. 
The size of hidden state in an LSTM cell is 256 and the word embedding layer has 64 dimensional output word vector.
They are initialized to be orthogonal each other and LSTM cells run up to the maximum size of the number of entities $T$ of video samples
A (shared) word projection layer generates a vocabulary distribution vector for each LSTM cell output. Then we calculate ${loss}_{word}$ by using standard softmax cross entropy for each output.
We examine whether this structure can show significant result or not.

\textbf{Guided LSTM with Stochastic Gating Mechanism}
We implemented a different structure of LSTMs guided by video-to-tag translation process. So the guided LSTM can act as a feature extractor for the connected classifier. For this experiment, we figure out which structural feedback variants of guided LSTM can perform better generalization than the base LSTM model. Here, we chose the baseline logistic model as our classifier following LSTMs.
All parameter settings are equal to the above base model, besides the additional one more LSTM step runs to generate hidden state to be input to the classifier. In addition, we try to apply binary cross entropy for ${loss}_{word}$ to be much faster learning convergence. We calculated additional ${loss}_{class}$ by using binary cross entropy for the final prediction, and optimize both ${loss}_{word}$ and ${loss}_{class}$

\textbf{Adding Batch Normalization layers into guided LSTM}
Since a batch normalization(BN) layer is powerful, it has become a trend to add BN layers to every layer in the structure.
However, our stochastic gating mechanism can distort the distribution of input word vectors. So we attempted to add BN layers gradually.
We firstly add a BN layer before each loss calculation. This preserves LSTM structure itself. Seconldy, we try to upgrade LSTM layer to BNLSTM layer. Lastly, we exploit both additions to figure out the performance improvement.

\textbf{Extention to other classifiers}
The above all models cooperate with logistic classifier model. To show our model can be a collaborative feature extractor with other classifiers, we reconnected our model to MoE model. For the further work, we show if our model can be upgraded as we use more competitive classifiers.
\section{Results and Discussion}
\begin{table}
\begin{center}
\begin{tabular}{|l|c|c|c|}
\hline
Model & Hit@1 & PERR & GAP \\
\hline\hline
Logistic Model & 82.5 & 69.1 & 75.9 \\
Mixture of Experts(MoE) & \textbf{83.9} & \textbf{70.7} & \textbf{78.0} \\
\hline\hline
ours & & & \\
\hline
max pooling & 80.9 & 66.5 & 73.0 \\
guided ($\beta=1.0$) & 81.4 & 67.5 & 74.4 \\
guided ($\beta=0.5$) & 81.8 & 68.0 & 75.1 \\
guided ($\beta=0.0$) & \textbf{82.5} & \textbf{68.9} & \textbf{76.3} \\
\hline
guided ($\beta=0.5$) & & & \\
LSTM + BN layer & 83.0 & 69.4 & 76.9 \\
BNLSTM & \textbf{83.9} & \textbf{70.8} & \textbf{78.3} \\
BNLSTM + BN layer & 83.6 & 70.2 & 77.9 \\
\hline
guided ($\beta=0.0$) & & & \\
LSTM + BN layer & 83.4 & 69.6 & 77.4 \\
BNLSTM & \textbf{84.3} & \textbf{71.2} & \textbf{78.8} \\
\hline
BNLSTM + MoE & \textbf{84.5} & \textbf{71.5} & \textbf{79.1} \\
\hline
\end{tabular}
\end{center}
\caption{Validation results of our models (100k iterations). All values in this table are averaged results and reported in percentage(\%).}
\end{table}

\textbf{Base and guided LSTM models}
We obtained two significant results from the first experiment: First, regardless of $\beta$, guided LSTM structure can perform better
than base model(GAP 73.0\%). Guidance process can make LSTM to learn dynamics which can generate the final hidden state tracked by label
entities. Second, ground truth label injection ($\beta=1$, 74.4\%) is prevalently used in neural machine translation or image description structures,
validation results show that metrics can be improved by decreasing $\beta$($\beta$=0.5, 75.1\%, $\beta$=0.0, 76.3\%). As described earlier, we can check that $\gamma_0 > \gamma(\beta)$.
In addition, since the activatio function is softmax after the word projection structure, it becomes a bottleneck of the base model.
The learning speed of guided LSTM, however, is increased by using individual binary cross entropy for each word projection, which get rid of the bottleneck.

\textbf{Batch Normalized LSTM and Extention with MoE}
We performed the second experiments for two different ground truth label injection probability ($\beta=0.5, 0.0$). As reported in Table 1., guided LSTM model has higher evaluation metrics by just adding a BN layer between guided LSTM and logistic classifier (76.9\%, 77.4\%). In addition, BNLSTM with logistic classifier (78.3\%) shows higher performance than that of the case of adding a BN layer (77.9\%). We had guessed that adding both modifications into the structure could perform better, but validation results of it don't show better results. We concluded that this may be related to stabilization of BN layers because each BN layer has its own population mean and variance which are accumulated by batch mean and variance values with exponential decaying algorithm and the result varies with this decaying factors. Above all, we got the higher results(78.8\%) with $\beta=0.0$ than base mixture model(78.0\%). In the final experiment, we obtained the possibility of extentions of our model by updating the highest value with different classifier(79.1\%)

\section{Conclusion}
In this paper, we proposed a method to use LSTMs as a feature extractor for multi-label video classification and investigated how to improve LSTM performance through batch normalization.
For better generalization, we found out that stochastic gating mechanism with $\beta=0.0$ has shown better validation results than $\beta>0.0$. it means it is better to use feedback loop from the previous LSTM cell in both training and inference phase. In addition, batch normalization layer improved the performance, but it requires careful consideration about which parts of the structure are attached by BN layer.
Last, mean pooling is known to be an effective aggregation method, but the ordered relation information that can be seen at frame-level may disappear by mean pooling. Therefore, it may be difficult to classify the same videos even if they have different meanings depending on the order of the frames. The transfer learning is performed by using the given CNN features from DB in this paper. To deal with frame-level features directly, we can put an LSTM encoder instead of mean pooling layer. In addition it can be a possible way to put an attention layer between encoder and decoer LSTMs for boosting up the overall metrics.

\section{Acknowledgements}

We thank Dr. Joonoo Kim in Mobile Communications Business of Samsung Electronics, who supported us to perform research and publish this work as a project leader, and Dr. Sundo Choi at Samsung Advanced Institute of Technology for his kind advice and helful discussion.

{\small
\bibliographystyle{ieee}
\bibliography{egpaper_for_review}
}

\end{document}